\documentclass{article}

\usepackage[final]{neurips_2021}

\usepackage{hyperref}       
\usepackage{url}            
\usepackage{booktabs}       
\usepackage{amsfonts}       
\usepackage{nicefrac}       
\usepackage{microtype}      
\usepackage{xcolor}         

\usepackage{epsfig}
\usepackage{graphicx}

\title{Fully convolutional Siamese neural networks for buildings damage assessment from satellite images}

\author{%
  Eugene Khvedchenya \\
  Computer vision research engineer\\
  Pi{\~n}ata Farms\\
  Odessa, Ukraine \\
  \texttt{ekhvedchenya@gmail.com} \\
   \And
   Tatiana Gabruseva\\
  Independent researcher \\
  Cork, Ireland \\
   \texttt{tatigabru@gmail.com} \\
}

\begin{document}

\maketitle

\begin{abstract}
Damage assessment after natural disasters is needed to distribute aid and forces to recovery from damage dealt optimally. This process involves acquiring satellite imagery for the region of interest, localization of buildings, and classification of the amount of damage caused by nature or urban factors to buildings. In case of natural disasters, this means processing many square kilometers of the area to judge whether a particular building had suffered from the damaging factors.

In this work, we develop a computational approach for an automated comparison of the same region's satellite images before and after the disaster, and classify different levels of damage in buildings. Our solution is based on Siamese neural networks with encoder-decoder architecture. We include an extensive ablation study and compare different encoders, decoders, loss functions, augmentations, and several methods to combine two images. The solution achieved one of the best results in the Computer Vision for Building Damage Assessment competition.
\end{abstract}

\section{Application Context}
Natural disasters require quick and accurate situational information for an effective response. Before responders can act in the affected area, they need to know the locations and severity of damage as soon as possible. Damage assessment involves acquiring satellite imagery for the region of interest, localization of buildings, and classification of damage degree caused to the buildings. High-resolution imagery is required to see the details of specific damage conditions. Current response strategies require in-person damage assessments within 24-48 hours of a disaster. The large areas affected combined with the vast numbers of pixels representing those areas make it laborious for analysts to search and evaluate damages in a disaster area. Therefore, an automated damage assessment of buildings after natural disasters would greatly help distribute humanitarian forces optimally. Here, we present the solution of the xView2 Computer Vision for Building Damage Assessment Challenge, hosted at https://www.xview2.org/. Our algorithm locates and compares buildings on two corresponding images before and after the disaster and classifies their damage level.

Deep learning approaches are widely used for the automatic extraction of information from satellite images~\cite{Solovyev2020, ternausnet,Buslaev_2018,Seferbekov_2018}. Siamese networks have shown great success in finding similar/dissimilar image pairs~\cite{Dey2017,Chicco2020}. Using a shared backbone to extract embeddings from two images, they can effectively learn discriminative features in the fully-connected layer. Siamese networks with encoder-decoder architecture were proposed for the change detection in~\cite{Caye2018}. Our solution is based on an end-to-end architecture that allows simultaneously locate objects on a pair of images and compare the difference between them. We describe practical techniques that help to improve the model's performance. We study various encoders, decoders, loss functions, augmentations, and different methods to combine input images. The source code is publicly available at~\cite{xview2} under the MIT license.

\section{Problem statement}
\label{sec:problem}
In the challenge, given a pair of images before ("pre") and after ("post") a natural disaster, solutions should output a semantic mask that corresponds to the "pre" image and classifies each pixel in one of the classes: 0 - "no building", 1 - "building, no damage", 2 - "building, minor damage", 3 - "building, major damage", and 4 - "building, destroyed". The details on the four-level damage scale are given in~\cite{RitwikGupta2019}.

A target objective in the challenge was a weighted sum of $F_1$ scores for building localization, $F_{1loc}$, and damage classification, $F_{1class}$ tasks, defined as follows:
\begin{equation}
Score = 0.3 * F_{1loc} + 0.7* F_{1class}.
\end{equation}

\section{Dataset}
The dataset xBD~\cite{RitwikGupta2019} contained $18336$ high-resolution satellite images, covering a diverse set of disasters with over $850,000$ buildings across over $45,000$ km2 of imagery. The dataset contained pairs of images of the same region before ("pre") and after ("post") a natural disaster and included data from $15$ countries and $6$ types of disasters. All images had a resolution of 1024 x 1024 pixels. The dataset was highly unbalanced in terms of damage degree, area of imagery, and polygons per disaster event. The distribution of damage classes was highly skewed towards "no damage"\cite{RitwikGupta2019}.

\begin{figure}[htb]
\centering
    \includegraphics[width=0.95\linewidth]{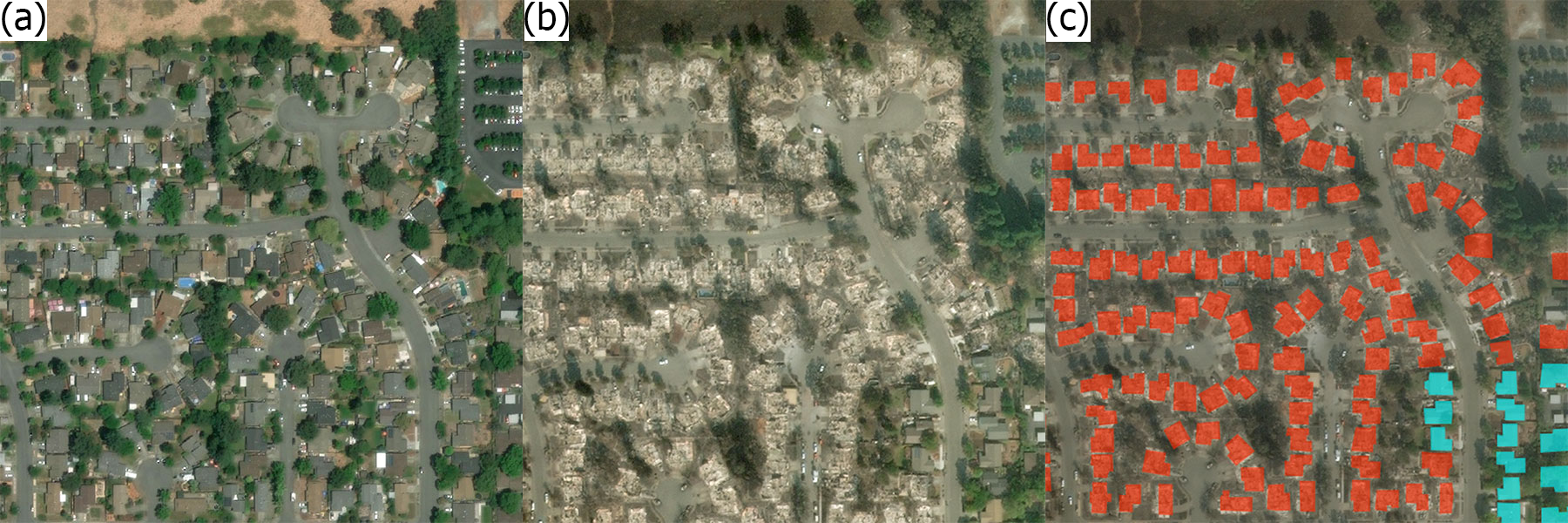}
   \caption{Examples of the satellite imagery collected before (a) and after (b) the disaster and the corresponding ground truth mask (c). The color of the buildings represents the damage degree, with totally destroyed buildings in red and undestroyed in blue.}
\label{fig:eda}
\end{figure}

The images were manually annotated with polygons. Each polygon represented a building/structure and was labeled with a degree of damage that occurred to it during the disaster, scaled from 1 to 4. The main objective of the challenge was to build a solution capable of processing a pair of satellite images of the same region before and after a disaster and assessing buildings' level of damage. Figure~\ref{fig:eda} shows an example of the satellite images and the corresponding ground truth mask.

\section{Methodology}
The goal in the given task is two-fold: a) detecting buildings in the "pre-disaster" image, b) evaluate each building's damage concerning its counterpart in the "post-disaster" image. One can solve this problem by performing semantic segmentation using pairs of input images.

For training, we generated masks with $1024$ x $1024$ resolution that contained $5$ classes with values: 0 - "no building", 1 - "building, no damage", 2 - "building, minor damage", 3 - "building, major damage", and 4 - "building, destroyed". The models were trained to predict a multi-class segmentation mask with five classes. All models were trained on crops of $512$ x $512$ px size and validated on full-resolution images of $1024$ x $1024$ px. For validation, the training set was split into five folds using multi-label stratification based on the amount and severity of damaged buildings. Five-folds cross-validation was used to find optimal model architectures and tune hyperparameters.

\subsection{Model architecture}
There is a wide range of deep learning architectures for semantic segmentation~\cite{Lin2016,Ronneberger2015,PSPNet}. Most of them follow the encoder-decoder principle. The encoder extracts representative feature maps of different spatial dimensions, and the decoder reconstructs a full-resolution semantic mask.

Unlike typical image segmentation tasks, this problem needs two images to produce predictions. There are several ways to input two images into the model. The most straightforward approach is to concatenate "pre-disaster" and "post-disaster" images from the pair into a single tensor with six channels and use it as an input. Another approach is to train the same backbone using both input images simultaneously. Siamese networks have shown great success in finding similar/dissimilar image pairs~\cite{Dey2017,Chicco2020,Caye2018}. Using a shared backbone to extract embeddings from two images, they can effectively learn discriminative features in the fully connected layer.

The Siamese network architecture for this task was designed as follows:
\begin{itemize}
\item Using a shared encoder extract feature maps of strides [2, 4, 8, 16, 32] for "pre-disaster" and "post-disaster" images.
\item Concatenate/subtract feature maps for each stride together along channels dimension.
\item Feed concatenated feature maps to the decoder (we studied Unet and FPN decoders).
\end{itemize}

\begin{figure}[htb]
\centering
   \includegraphics[width=0.6\linewidth]{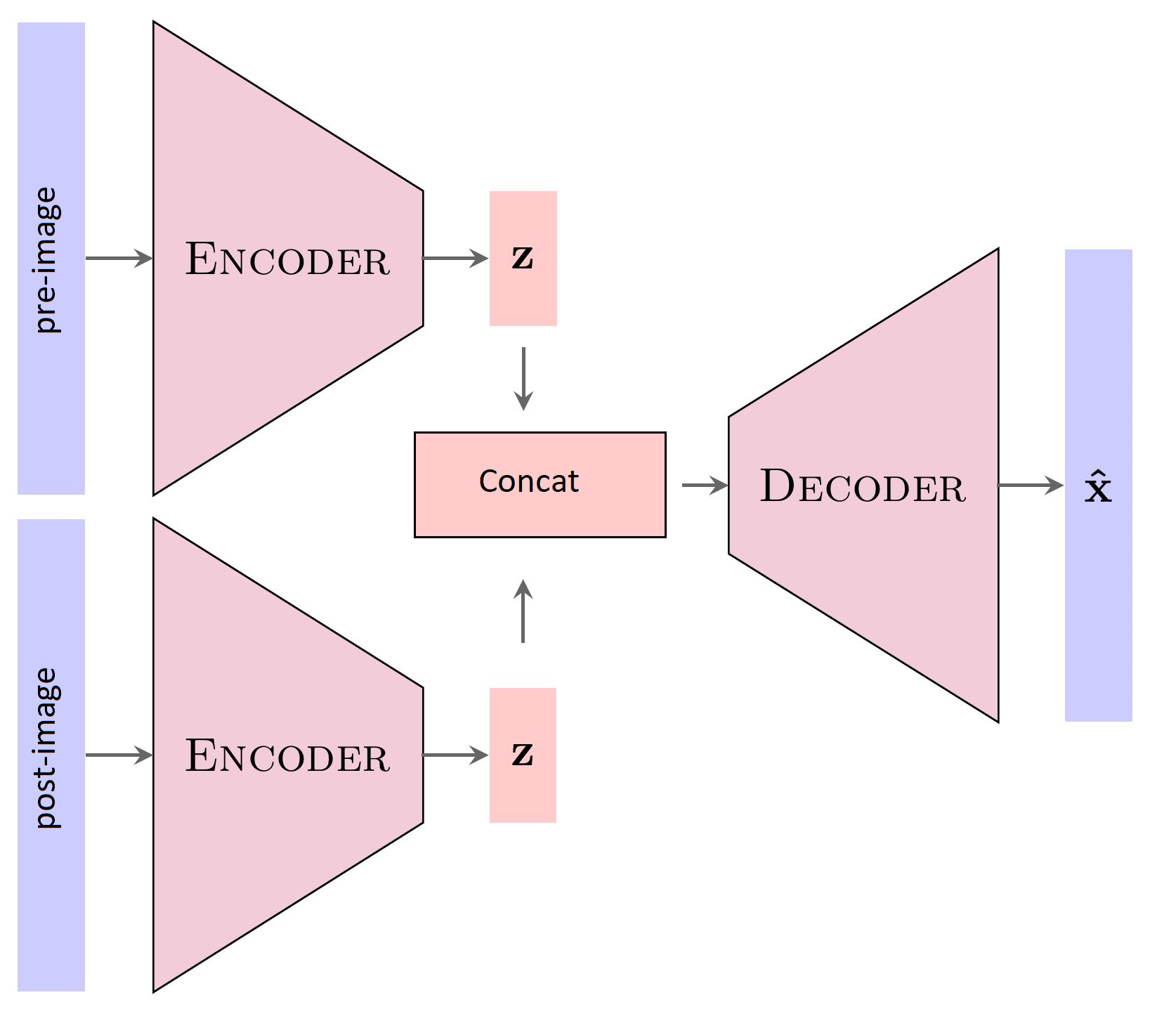}
   \caption{Schematic architecture of the hybrid Siamese encoder-decoder architecture.}
\label{fig:net}
\end{figure}
This hybrid Siamese architecture is shown schematically in Figure~\ref{fig:net}.

\subsection{Images pre-processing and augmentations}
The original images were cropped and resized to 512 x 512 px resolution for training. The validation was performed on the full-resolution images, 1024 x 1024 px.
For augmenting images, we used an open-source library Albumentations~\cite{albu}.
The following image augmentations were applied during training phase:
\begin{itemize}
\item \textbf{Spatial transformation, applied to only "post" images:} random rotation up to 3 degree, random shift by (-10,+10) pixels, random scale in the range -2\% .. +2\%
\item \textbf{Spatial transformation, applied to both "pre" and "post" images:}
    crop 512x512 patches for same region, random scale varying from 80\% to 120\%,
	horizontal flip, rotation by 90 degrees, transpose,
    random grid shuffle,
    mask dropout
\item \textbf{Color augmentations, applied independently to "pre" and "post" images:}
    random brightness, contrast and gamma change,
    changes in HSV or RGB colorspace
\end{itemize}
Heavy spatial augmentations where applied in the same manner to both "pre" and "post" images to ensure consistency in their crops. Color augmentations, on the other side, were applied independently.

\section{Results: Ablation study}

\subsection{Combining input images: basic model architecture}
We compare Siamese networks to the classical segmentation approach, based on concatenation of "pre-disaster" and "post-disaster" into a 6-channel tensor and sending it through an encoder-decoder model. The results are shown in Table 2 for various ResNet~\cite{resnet} encoders, Unet decoder, and identical training parameters. We show localization, classification and weighted $F_1$ scores obtained for different models' architectures on a validation set. The Siamese architectures outperform by far the classical encoder-decoder architecture, especially for the damage classification task. The concatenation of features in the Siamese network outperformed subtraction for most of the encoders, although the difference is relatively small.

\begin{table}[htb]
\caption{Localization, damage and weighted $F_1$ scores obtained for different models' architectures on a validation set}
\centering
\begin{tabular}{lcccc}
\toprule
Approach & Encoder/Decoder & $F_{1loc}$ & $F_{1class}$ & $F_1$ \\
\midrule
 concatenation & ResNet18/Unet & 0.8649 & 0.6784 & 0.7344 \\
 concatenation & ResNet34/Unet & 0.8475 & 0.6098 & 0.7211 \\
 concatenation & ResNet50/Unet & 0.8707 & 0.6833 & 0.7395 \\
\midrule
Siamese, concat & ResNet18/Unet & 0.8638 & 0.6831 & 0.7373 \\
Siamese, concat & ResNet34/Unet & 0.8720 & 0.7249 & 0.7690 \\
Siamese, concat & ResNet50/Unet & 0.8726 & 0.7218 & 0.7670 \\
\midrule
Siamese, subtract & ResNet18/Unet & 0.8618 & 0.7255 & 0.7664 \\
Siamese, subtract & ResNet34/Unet & 0.8616 & 0.7166 & 0.7601 \\
Siamese, subtract & ResNet50/Unet & 0.8664 & 0.7188 & 0.7631 \\
\bottomrule
\end{tabular}
\label{table:siamese}
\end{table}

\subsection{Effect of augmentations and weighted entropy loss}
\label{augs}
We ran an ablation study using a Siamese architecture with ResNet-34 encoder and Unet decoder to measure the influence of image augmentations on the model performance. The results are presented in Table 2. Using both spatial and color augmentations helped improve the model accuracy, especially for the damage classification task, compared to the setting without transforms or for color transforms only. Using strong transformations did not bring any further improvements in the model performance. A default choice for loss function for semantic segmentation is cross-entropy (CE) loss. In the damage detection task, a critical problem is overcoming classes imbalance, as the majority of buildings are not affected or affected less severely. To address this challenge, we propose a weighted CE loss. We assigned weights of $1.0$ to the most common "not building" and "no damage" classes and weights of $3.0$ to less popular "minor", "major" and "destroyed" classes. Such weighting of under-represented classes helped to increase $F_1$ score on those classes. We found that a CE loss with class weights worked better for this challenge as it provides compensation for the class imbalance.

\begin{table*}[htb]
\caption{Localization and  damage classification $F_1$ scores, and weighted $F_1$ score achieved for Siamese ResNet-34/Unet architecture for different losses and augmentations}
\centering
\begin{tabular}{lccccc}
\toprule
Model & Loss & Augmentations & $F_{1loc}$ & $F_{1class}$ & Score\\
\midrule
ResNet34/Unet & CE &    Medium & 0.8765  & 0.7096 & 0.7597 \\
ResNet34/Unet & Weighted CE & Medium & 0.8720   & 0.7249 & \textbf{0.7690} \\
ResNet34/Unet & Weighted CE & None & 0.8704 & 0.7068 & 0.7559 \\
ResNet34/Unet & Weighted CE & Color only&  0.8716  & 0.7061 & 0.7558 \\
ResNet34/Unet & Weighted CE & Hard &  0.8712  & 0.7246 & 0.7686 \\
\bottomrule
\end{tabular}
\label{table:lossaug}
\end{table*}

\subsection{Different encoders and decoders}
\label{encoders-decoders}
We used feature extractors pre-trained on ImageNet dataset~\cite{cadene}. A number of different encoder architectures has been studied: ResNet-18, -34, -50, -101~\cite{resnet}, DenseNet-169, -201~\cite{densenet}, SE-ResNext-50~\cite{Hu_2018_CVPR}, Inception-v4~\cite{inc}, and EfficientNet~\cite{effnet}. Table 3 shows validation scores for various encoders and decoders architectures. The lightest encoders, like ResNet-18, produced significantly lower results for both building localization and damage classification, showing underfitting for this problem. The models with heavier encoders generally achieved better results for the same decoder type. The SE-ResNext-50 encoder architecture demonstrated the best performance on this dataset both for localization and classification, achieving a weighted $F_1$ score of $0.78$ for a single fold model. DenseNets also gave close results.

\begin{table}[htb]
\caption{Scores for buildings localization, $F_{1loc}$, damage classification, $F_{1class}$, and weighted $F_1$ score, achieved with various encoder-decoder models' architectures on a validation set.}
\centering
\begin{tabular}{lcccc}
\toprule
Encoder & Decoder & $F_{1loc}$ & $F_{1class}$ & Score\\
\midrule
ResNet18 & Unet & 0.8638 & 0.6831 & 0.7373 \\
ResNet34 & Unet	& 0.8720 & 0.7249 & 0.7690 \\
ResNet50 & Unet & 0.8726 & 0.7218 & 0.7670 \\
ResNet101& Unet & 0.8738 & 0.7225 & 0.7679 \\
DenseNet169 & Unet & 0.8740& 0.7293 & 0.7727\\
Se-ResNext50 & Unet & \textbf{0.8797} & \textbf{0.7338} & \textbf{0.7776}\\
\midrule
ResNet101 & FPN & 0.8710 & 0.7143 & 0.7613\\
Inception-v4 & FPN & 0.8332 & 0.7199 & 0.7539\\
Efficient-b4 & FPN &	0.8605 & 0.7213& 0.7631\\
\bottomrule
\end{tabular}
\label{table:results}
\end{table}

The Unet decoder produced slightly better results than FPN for the same encoders. The difference in performance for the Unet or FPN was relatively small. The choice on the pre-trained encoder had higher influence on the overall model performance, especially, for the classification task.

\section{Conclusions}
In this work, we develop a computational approach for automated building localization and their damage assessment from a pair of satellite images.
We consider several methods to combine the input images or their feature maps. The Siamese architecture outperformed by far the classical approach with concatenated input images and achieved one of the best results in the Computer Vision for Building Damage Assessment competition. Concatenating feature maps before the decoder part performed slightly better than their substraction. We tested different decoders and encoders and introduced a number improvements were implemented that helped to increase the performance of the model. The extensive ablation study has shown the effect of these modifications on the model's performance.
The source code is publicly available at~\cite{xview2}.



\medskip

{\small
\bibliographystyle{plainnat}
\bibliography{xviewbib}
}
\appendix

\section{Appendix A. Images augmentations}
\begin{figure}
  \centering
  \includegraphics[width=0.99\linewidth]{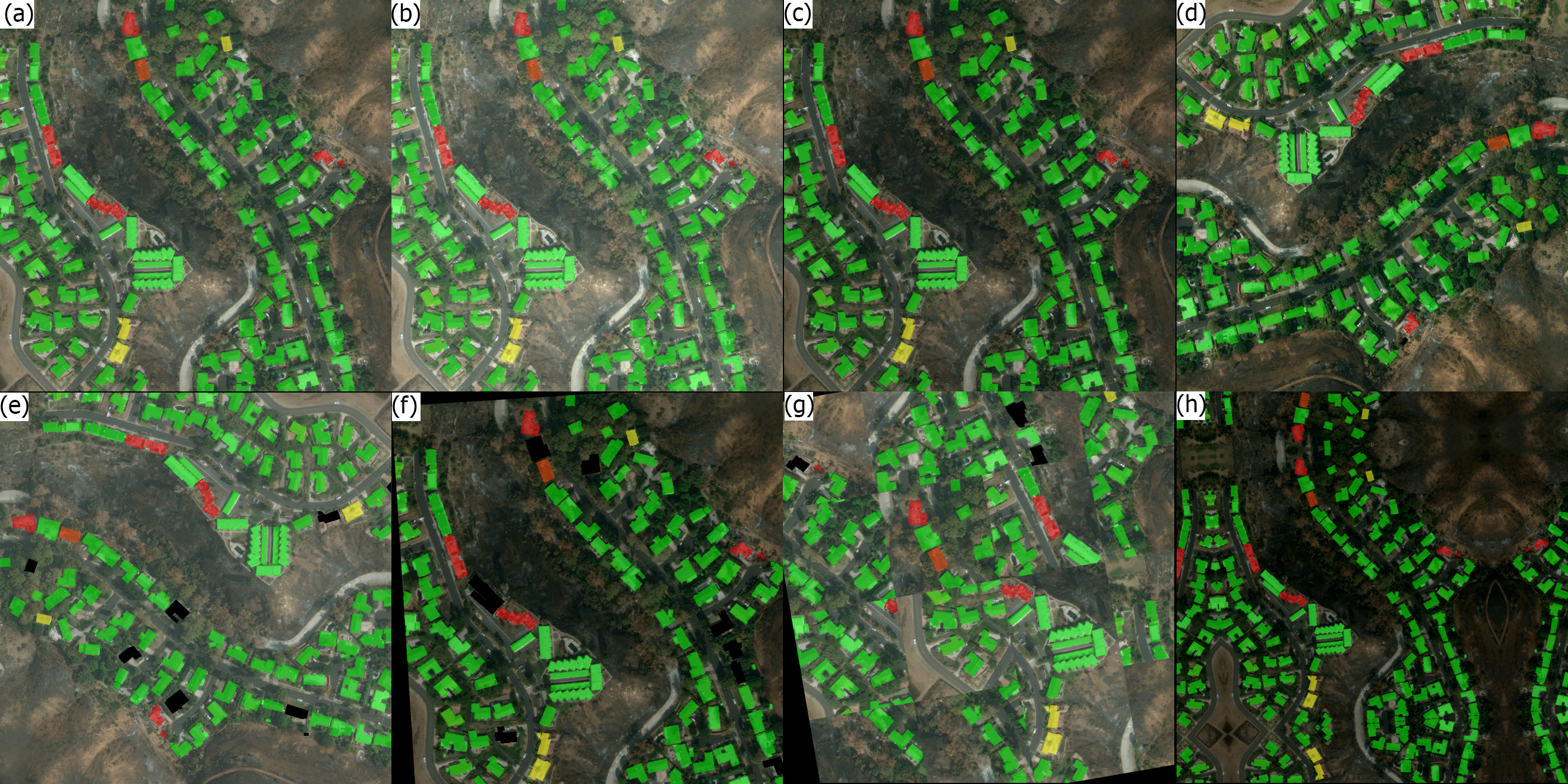}
   \caption{A satellite imagery sample with annotations (a) and its transformations (b-h).}
\label{fig:aug}
\end{figure}

Image augmentation played a crucial part in the training pipeline. Augmentations helped prevent model overfitting, and in fact, was a critical component that increased models' performance, especially for the pairs with significant displacement of "pre" and "post" images. For augmenting images, we used an open-source library Albumentations~\cite{albu} that allows creating augmented images on-the-fly during training, augments both images and corresponding masks simultaneously, and has a diverse number of supported transforms.
An example of an image sample and its augmented variants is shown in Figure~\ref{fig:aug}.

\section{Appendix B. Training procedure}
Deep learning models were trained using a PyTorch framework~\cite{pytorch} and a Catalyst library~\cite{catalyst}. PyTorch is an efficient and flexible framework that allows the construction of custom models' architectures and works well in this problem due to the customized input format for the models. Catalyst is a high-level framework above PyTorch that supports mixed-precision training, logging, and multi-GPUs training out of the box. For designing models' architecture, we used a pytorch-toolbelt library~\cite{PyTorchToolbelt} that allows plug-and-play change of encoders, decoders, and their parameters and provides a flexible way of creating building blocks of the model.

The training hyper-parameters for all experiments are summarized in Table 4. The batch size varied from $32$ to $64$ samples, depending on model architecture. We used the RAdam optimizer~\cite{Liu2019} with $1e-3$ learning rate. The full list of parameters can be found in the source code repository~\cite{xview2}.
\begin{table}[htb]
\caption{Training hyper-parameters}
  \centering
  \begin{tabular}{lc}
  \toprule
  Parameter & Description \\
  \midrule
      Optimizer & RAdam  \\
      Batch size & 32 .. 64 \\
      Weight decay & 1e-5 \\
      Learning rate (Lr) & 1e-3 \\
      Lr scheduler & Cosine decay \\
      Minimum Lr & 1e-6 \\
      Epochs & 100 \\
      Train, crop size & 512 x 512 \\
      Validation, image size & 1024 x 1024 \\
  \bottomrule
  \end{tabular}
  \label{table:params}
\end{table}

All models were trained on random-sized crops of $512$ x $512$ px size and validated on full-resolution images of $1024$ x $1024$ px. The best model checkpoints were chosen based on the evaluation metric. The training was performed on 4 NVidia GeForce $1080Ti$ GPUs and a p3.8xlarge AWS Instance, equipped with 4 NVidia Tesla V100 GPUs.

For validation, the training set was split into $5$ folds. As a splitting algorithm, we used multi-label stratification based on the amount of non-damaged, minor-damaged, major-damaged, and destroyed buildings in each image and event. Such a split guarantees that each fold will contain the same distribution of damage classes with respect to event types. The alternative approach is to use a group K-fold to ensure images from one event belong only to the same fold. Five-folds cross-validation was used to find optimal model architectures and tune hyperparameters.

\section{Appendix C. Ensembling}
We computed the weighted average of class probabilities of all models' predictions and then applied argmax to get class labels for each pixel in the final prediction.  We computed weighting coefficients for each class by optimizing the weighted $F_1$ score on out-of-fold predictions. The tuned class weights for all models were [0.5, 1.1, 1.1, 1.1, 1.1], which improved weighted $F_1$ score at $+0.01-0.02$ for single models, and by $+0.04$ $F_1$ for the entire ensemble.

The final ensemble included 13 models from 5 folds, selected so that there were at least 2 models in the ensemble for each fold. We used different encoder backbones and Unet or FPN decoders types. The ensemble achieves a weighted $F_1$ score of $0.803$ on the competition hold-out test.

\end{document}